\documentclass{midl} 
\usepackage{caption} 
\usepackage{float}
\usepackage{wrapfig}
\usepackage{diagbox}  
\usepackage{multirow}
\usepackage{booktabs}
\usepackage{dsfont}
\usepackage{bbm}
\usepackage[british]{babel}
\usepackage{microtype}
\usepackage[title]{appendix}

\newcommand{\simclr}{\mbox{SimCLR}}
\newcommand{\tsne}{$t$\babelhyphen{nobreak}SNE}
\newcommand{\tsimcne}{$t$\babelhyphen{nobreak}SimCNE}

\jmlryear{2024}

\title[Self-supervised Visualisation of Medical Image Datasets]{Self-supervised Visualisation of Medical Image Datasets}

\midlauthor{%
 \Name{Ifeoma Veronica Nwabufo\nametag{$^{1,2}$}} \Email{ifeoma-veronica.nwabufo@uni-tuebingen.de}\\
 \Name{Jan Niklas B{\"o}hm\nametag{$^{1,2}$}} \Email{jan-niklas.boehm@uni-tuebingen.de}\\
 \Name{Philipp Berens\nametag{$^{1,2}$}} \Email{philipp.berens@uni-tuebingen.de}\\
 \Name{Dmitry Kobak\nametag{$^{1,2,3}$}} \Email{dmitry.kobak@uni-tuebingen.de}\\
 \AND
 \addr $^{1}$ Hertie Institute for AI in Brain Health, University of T{\"u}bingen, Germany\\
 \addr $^{2}$ Tübingen AI Center, University of T{\"u}bingen, Germany\\
 \addr $^{3}$ IWR, Heidelberg University, Germany
}

\begin{document}
\maketitle

\begin{abstract}
Self-supervised learning methods based on data augmentations, such as SimCLR, BYOL, or DINO, allow obtaining semantically meaningful representations of image datasets and are widely used prior to supervised fine-tuning. A recent self-supervised learning method, $t$-SimCNE, uses contrastive learning to directly train a 2D representation suitable for visualisation. When applied to natural image datasets, $t$-SimCNE yields 2D visualisations with semantically meaningful clusters. In this work, we used $t$-SimCNE to visualise medical image datasets, including examples from dermatology, histology, and blood microscopy. We found that increasing the set of data augmentations to include arbitrary rotations improved the results in terms of class separability, compared to data augmentations used for natural images. Our 2D representations show medically relevant structures and can be used to aid data exploration and annotation, improving on common approaches for data visualisation.
\end{abstract}


\section{Introduction}

Medical image datasets have been quickly growing in size and complexity \cite{litjens2017survey, topol2019high, zhou2021review}. Whereas medical professionals can analyse, annotate, and classify individual images, tasks involving large batches of images, ranging from data curation and quality control to exploratory analysis, 
remain challenging.

Self-supervised learning (SSL) has recently emerged in computer vision as the dominant paradigm for learning image representations suitable for downstream tasks \cite{balestriero2023cookbook}, and it has increasingly been adopted in medical imaging \cite{huang2023self}. In \textit{contrastive learning} methods, such as SimCLR \cite{chen2020simple}, BYOL \citep{grill2020bootstrap}, or DINO \cite{caron2021emerging}, data augmentation is used to generate different \textit{views} of each image, and a deep network is trained to keep these views close together in the representation space. However, the learned representations are typically high-dimensional.

\begin{figure}[t]
    \includegraphics{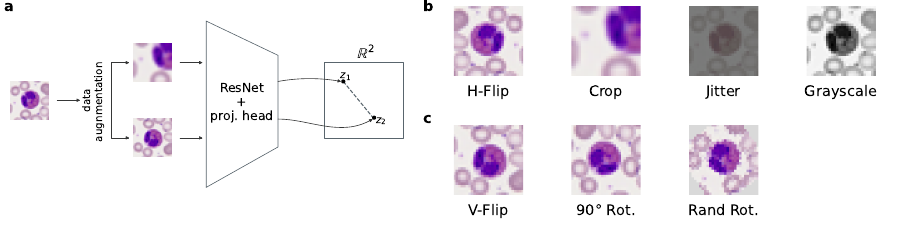}
    \caption{\textbf{(a)} In $t$-SimCNE, the network is trained to map two random augmentations of an input image to close locations in the 2D output space. \textbf{(b)} Augmentations used for natural images in $t$-SimCNE. \textbf{(c)} Additional augmentations suggested here for medical images.}
    \label{fig:augm-isolation}
\end{figure}

Recently, \citet{boehm2023unsupervised} suggested a self-supervised contrastive method, called \tsimcne, for 2D visualisation of image datasets. Using natural image datasets, the authors demonstrated that \tsimcne{} obtains semantically meaningful visualisations, representing rich cluster structure and highlighting artefacts in the data. Their methods clearly outperformed existing 2D embedding methods like $t$-SNE \cite{van2008visualizing} and UMAP \cite{mcinnes2020umap} for natural image data.

Here we apply \tsimcne{} to several medical microscopy datasets, and demonstrate that it yields medically relevant visualisations,  outperforming $t$-SNE visualisations of pretrained networks. Furthermore, we show that the results improve when using rotational data augmentations (Figure~\ref{fig:augm-isolation}) informed by the rotational invariance of microscropy images. Our code is available at \url{https://github.com/berenslab/medical-t-simcne}.

\section{Related work}

Contrastive learning methods have been widely applied to medical image datasets \citep[for a review, see][]{huang2023self} but usually as pre-training for downstream tasks such as classification or segmentation. Some recent works visualised high-dimensional SSL representations; e.g. \citet{cisternino2023self} used UMAP of DINO to visualise histopathology data. In contrast, our focus is on self-supervised visualisations trained end-to-end.

Contrastive learning relies on data augmentations to create several views of each image, and the choice of data augmentations plays a crucial role in methods' success \cite{tian2020makes}. 
A large number of works explored data augmentations for medical images in a supervised setting \cite[reviewed by][]{chlap2021review,goceri2023medical}. In the self-supervised context, \citet{van2023exploring} studied the effect of augmentations on the representation of X-ray images. For histopathology images, \citet{kang2023benchmarking} advocated for using rotations and vertical flips, as well as staining-informed color transformations, while some other works used neighbouring patches as positive pairs \cite{li2021dual, wang2021transpath}.

\section{Background: SimCLR and $t$-SimCNE}
\label{sec:background}

SimCLR \citep{chen2020simple} produces two augmentations for each image in a given mini-batch of size $b$, resulting in $2b$ augmented images. Each pair of augmentations forms a so-called \textit{positive pair}, whereas all other possible pairs in the mini-batch form \textit{negative pairs}. The model is trained to maximise the similarity between the positive pair elements while simultaneously minimising the similarity between the negative pair elements. 

An augmented image $x_i$ is passed through a ResNet \cite{he2016deep} \textit{backbone} to give the latent representation $h_i$, which is then passed through a fully-connected \textit{projection head} with one hidden layer to yield the final output $z_i$. SimCLR employs the InfoNCE loss function \cite{oord2018representation}, which for one positive pair $(i,j)$ can be written as
\begin{equation}
\label{simclr}
    \ell_{ij}=-\log \frac{\exp\big(\text{sim}(z_{i},z_{j})/\tau\big)}{\sum_{k \ne i}^{2b}\exp\big(\text{sim}(z_{i},z_{k})/\tau\big)}\,,
\end{equation}
where $\text{sim}(x,y)= x{^\top}y/\big(\|x\|\cdot\|y\|\big)$ is the cosine similarity and $\tau$ is a hyperparameter that was set to $1/2$ in \citet{chen2020simple}. Even though the loss function operates on $z_i$ (typically 128-dimensional), for downstream tasks, SimCLR uses the representations $h_i$ \cite{bordes2022guillotine}, typically at least 512-dimensional.

The idea of $t$-SimCNE \cite{boehm2023unsupervised} is to make the network output ($z_i$) two-dimensional so that it is directly suitable for data visualization. It does not make sense to apply the cosine similarity to this representation in $\mathbb R^2$ as it would effectively normalise the embeddings to lie on a one-dimensional circle. $t$-SimCNE replaces the exponential of the scaled cosine similarity with the Cauchy similarity used in $t$-SNE \citep{van2008visualizing}: $(1 + \|x-y\|^2)^{-1}$. The resulting loss function is
\begin{equation}
    \ell_{ij}=-\text{log} \frac{1}{1+\|z_i - z_j\|^{2}} + \text{log} \sum_{k \ne i}^{2b} \frac{1}{1+\|z_i - z_k\|^{2}}\,.
    \label{$t$-SimCNE_equation}
\end{equation}

\citet{boehm2023unsupervised} found that directly optimizing this loss is difficult, and suggested a three-stage process. The first stage (1000 epochs) used a 128-dimensional output which was then replaced with a 2D output and fine-tuned in the subsequent two stages (500 epochs). 

For their experiments on CIFAR datasets, the authors used a ResNet18 with a modified first layer kernel size of $3\times 3$, and a projection head with hidden layer size of~1024 (Figure~\ref{fig:augm-isolation}a).

\section{Experimental setup}
\label{sec:methods}

\begin{table}[t]
\caption{Used datasets.}
\centering
\small
\begin{tabular}{lcrrl}
\toprule
\textbf{Dataset}  & \textbf{Image dim.} &\textbf{Sample size}  & \textbf{Classes} & \textbf{Reference}\\ \midrule
Leukemia & $28 \times 28$ & $18\,365$ & $7$ &\citet{matek2019singlecell} \\ 
Blood\scshape{mnist} & $28 \times 28$ & $17\,092$ & $8$&\citet{medmnistv2}\\ 
Derma\scshape{mnist} & $28 \times 28$ & $10\,015$ & $2$&\citet{medmnistv2}\\
Path\scshape{mnist} &  $28 \times 28$ & $107\,180$ & $9$&\citet{medmnistv2}\\
PCam16 & $96 \times 96$ & $327\,680$ & $2$&\citet{veeling2018rotation}\\ 
\bottomrule
\end{tabular}
\label{tab:dataset-summary}
\end{table}

\paragraph{Datasets}

We used five publicly available medical image datasets with sample sizes ranging from 10\,000 to over 300\,000 (Table~\ref{tab:dataset-summary}). Three datasets we took from the Med{\scshape mnist}v2 collection \citep{medmnistv2}, all consisting of $28 \times 28$ RGB images. Derma{\scshape mnist} is based on the HAM10000 dataset \cite{tschandl2018ham10000}, a collection of multi-source dermatoscopic images of common pigmented skin lesions. The images are categorised into 7 classes, which we reduced to binary labels: melanocytic nevi and other skin conditions. Blood{\scshape mnist} is based on a dataset of microscopy images of individual blood cells from healthy donors \cite{acevedo2020dataset}, categorised into $8$ classes corresponding to cell types.  Path{\scshape mnist} is based on a dataset of non-overlapping patches from colorectal cancer histology slides \cite{kather2019predicting}, categorized into $9$ classes corresponding to tissue types. The Leukemia dataset \cite{matek2019human} contains microscopy images of white blood cells taken from patients, some of which were diagnosed with acute myeloid leukemia. We resized $224 \times 224$ images to $28 \times 28$ and merged 9 rare classes ($<80$ cells) into one, obtaining 7 classes.

The Patch Camelyon16 (PCam16) dataset \cite{veeling2018rotation}, adapted from the Camelyon16 challenge \cite{bejnordi2017diagnostic}, consists of   $96 \times 96$ patches from breast cancer histology slides. There are two classes: metastases and non-metastases. A patch was classified as metastases if there was any amount of tumor tissue in its central $32 \times 32$ region.

\begin{figure}[t]
    \includegraphics[width=\textwidth]{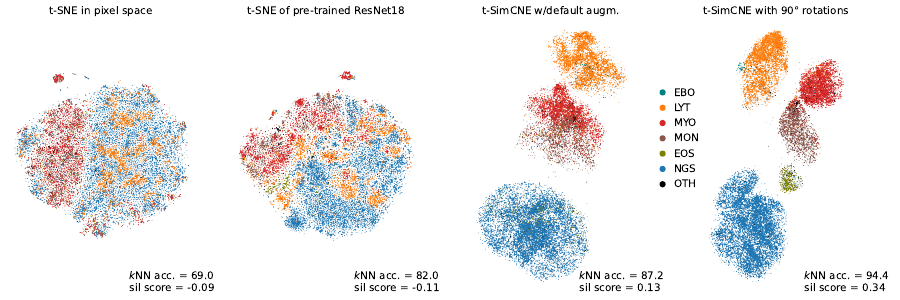}
    \caption{Visualisations of the Leukemia dataset. Small classes shown in black (`OTH' in the legend). $k$NN accuracy and silhouette scores shown in each panel. \textbf{(a)} $t$-SNE of the original images in the pixel space. \textbf{(b)} $t$-SNE of the 512-dimensional representation obtained via an ImageNet-pretrained ResNet18 network. \textbf{(c)} $t$-SimCNE using the same augmentations as in  \citet{boehm2023unsupervised}. \textbf{(d)} $t$-SimCNE using augmentations including 90° rotations and flips. Note that the EBO class is well separated here, despite only consisting of 78 images.}
    \label{fig:leukemia}
\end{figure}

\paragraph{Augmentations}

\Citet{boehm2023unsupervised} worked with natural images and used the same data augmentations as \citet{chen2020simple}: cropping, horizontal flipping, color jittering, and grayscaling (Figure~\ref{fig:augm-isolation}b). Here we used all of these augmentations with the same hyperparameters and probabilities (see Table~\ref{tab:ablations} for ablations). We reasoned that the semantics of microscopy or pathology images should be invariant to arbitrary rotations and arbitrary flips \citep{kang2023benchmarking}. For that reason we considered two additional sets of augmentations: (i) vertical flips and arbitrary 90° rotations; (ii) rotations by any arbitrary angle (Figure~\ref{fig:augm-isolation}c). In each case, all possible rotations were equally likely. When rotating an image by an angle that is not a multiple of 90°, the corners need to be filled in (Figure~\ref{fig:augm-isolation}c, right). For this we used the average color of all border pixels across all images in a given dataset. This color was dataset specific, but the same for all images in a dataset.

\paragraph{Architecture and training}

We used the original $t$-SimCNE implementation \cite{boehm2023unsupervised} with default parameters unless stated otherwise. For PCam16, we used the unmodified ResNet18  \cite{he2016deep} without the fully-connected layer. All networks were trained from scratch on an NVIDIA RTX A6000 GPU with the batch size of 1024, except for PCam16 where we had to reduce the batch size to 512 to fit it into GPU memory.

\paragraph{Baselines}

For comparison, we applied $t$-SNE to images in pixel space, in pretrained ResNet representation, and in SimCLR representation. The SimCLR models had the same architecture as $t$-SimCNE models but with 128D output and were trained with  SimCLR loss (Eq.~\ref{simclr}) for 1000 epochs. We then applied $t$-SNE to the 512-dimensional SimCLR representation before the projector head. We took ImageNet-pretrained ResNet18 and ResNet152 models from the PyTorch library \cite{Paszke_PyTorch_An_Imperative_2019}. To pass our images through these networks, we resized all images to $256\times 256$, center cropped to $224\times 224$, and normalized \citep[following][]{he2016deep}. The resulting representations had 512 and 2048 dimensions respectively. We used openTSNE $1.0.1$ \cite{Polivcar731877} with default settings to reduce to 2D. When doing $t$-SNE of the PCam16 data in pixel space, we first performed principal component analysis and only used the first $100$ PCs as input to $t$-SNE. 

\begin{figure}[t]
    \includegraphics{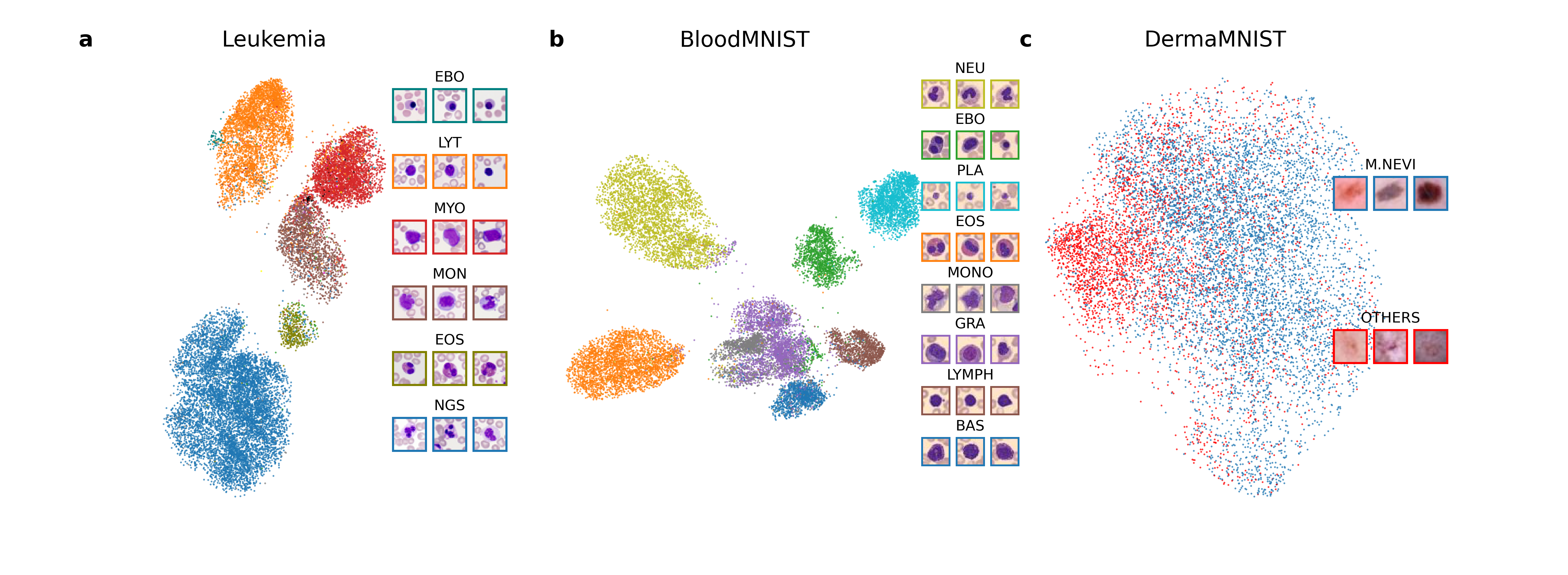}
    \caption{\textbf{(a)} $t$-SimCNE visualisation of the Leukemia dataset. Only a subset of classes is listed in the legend. \textbf{(b)} $t$-SimCNE visualisation of the Blood{\scshape mnist} dataset. \textbf{(c)} $t$-SimCNE visualisation of the Derma{\scshape {\scshape mnist}} dataset. In all three cases, we used augmentations including 90° rotations and vertical flips.}
    \label{fig:f3_toy_annotation}
\end{figure}

\paragraph{Evaluation}

We used two metrics to evaluate the quality of 2D embeddings, with classification and clustering being two possible downstream tasks: $k$NN classification accuracy \cite{scikit-learn} with $k = 15$ and a 9:1 training/test split, and silhouette score \citep{rousseeuw1987silhouettes}. For a single point $x$, the silhouette score $s\in[-1,1]$ is defined as $(b-w)/\max(w,b)$ where $w$ is the average distance between $x$ and points from the same class, and $b$ is the average distance between $x$ and points from the closest other class. The silhouette score of the entire embedding is the average $s$ across all points. These two measures are complementary:  The $k$NN accuracy measures how \textit{well} the classes are separated from each other, while the silhouette score measures how \textit{far} they are separated from each other.

\section{Results}

\begin{table}[t]
    \caption{\label{tab:accuracy} The $k$NN accuracy of 2D embeddings. Means $\pm$ standard deviations over three runs; PCam16 experiments had only one run due to its large size.} 
    \centering\footnotesize
    \begin{tabular}{clccccc}
        \toprule
        \multicolumn{2}{c}{\multirow{2}{*}{Method}}& \multicolumn{5}{c}{Dataset} \\
        \cmidrule{3-7}
        && Leukemia & Blood\scshape{mnist} & Derma\scshape{mnist} & Path\scshape{mnist} & PCam16 \\
        \midrule
        \multirow{3}{*}{\tsimcne} & def. augm. & $86.3 \pm 0.7\%$ & $90.4 \pm 0.3\%$ & $77.3 \pm 0.6\%$ & $97.2 \pm 0.2\%$ & $92.6 \%$ \\
        
        & + 90° rot. & $94.4 \pm 0.1\%$ & $93.0 \pm 0.3\%$ & $77.5 \pm 0.3\%$ & $98.0 \pm 0.0\%$ & $93.1\%$ \\
        
        & + rand. rot. & $95.1 \pm 0.2\%$ & $92.9 \pm 0.1\%$ & $80.1 \pm 0.7\%$ & $97.3 \pm 0.0\%$ & $90.8\%$ \\
        
        \midrule
        \multirow{3}{*}{\begin{tabular}{c}
             \tsne\ of  \\
             \simclr
        \end{tabular}} & def. augm. & $95.0 \pm 0.1\%$ & $94.0 \pm 0.1\%$ & $81.9 \pm 0.1\%$ & $98.1 \pm 0.0\%$ & $96.3\%$ \\
        & + 90° rot. & $95.9 \pm 0.1\%$ & $95.8 \pm 0.1\%$ & $80.8 \pm 0.6\%$ & $98.4 \pm 0.0\%$ & $96.4\%$ \\
        & + rand. rot. & $95.6 \pm 0.1\%$ & $95.4 \pm 0.1\%$ & $82.2 \pm 0.2\%$ & $97.9 \pm 0.1\%$ & $94.9\%$ \\
        \midrule
  
        \multirow{3}{*}{\tsne} & pixel space& $69.0\%$ & $73.2\%$ & $78.0\%$ & $56.9\%$ & $76.9\%$ \\
        
        &  ResNet18 & $82.0\%$ & $78.1\%$ & $81.9\%$ & $87.2\%$ & $86.7\%$ \\
        
        &  ResNet152 & $72.9\%$ & $72.9\%$ & $81.0\%$ & $88.8\%$ & $86.4\%$ \\
        \bottomrule
    \end{tabular}
\end{table}

\begin{table}[t]
   \caption{\label{tab:silhoutte_score} Silhouette scores (Section~\ref{sec:methods}) of 2D embeddings. Same format as in Table~\ref{tab:accuracy}.}
    \centering\footnotesize
    \begin{tabular}{clccccc}
        \toprule
        \multicolumn{2}{c}{\multirow{2}{*}{Method}}& \multicolumn{5}{c}{Dataset} \\
        \cmidrule{3-7}
        && Leukemia & Blood\scshape{{\scshape mnist}} & Derma\scshape{{\scshape mnist}} & Path\scshape{{\scshape mnist}} & PCam16 \\
        \midrule
        \multirow{3}{*}{\tsimcne} & def. augm. & $0.13 \pm 0.00$ & $0.40 \pm 0.00$ & $0.13 \pm 0.01$ & $0.45 \pm 0.02$& $0.04$ \\        
        & + 90° rot. & $0.33 \pm 0.01$& $0.44 \pm 0.03$&$0.11 \pm 0.00$&$0.48 \pm 0.06$&$0.05$\\        
        & + rand. rot. &$0.52 \pm 0.02$ &$0.50 \pm 0.01$&$0.13 \pm 0.06$&$0.41 \pm 0.03$&$0.05$\\
        \midrule
        \multirow{3}{*}{\begin{tabular}{c}
             \tsne\ of  \\
             \simclr
        \end{tabular}} & def. augm. &$0.21 \pm 0.01$ &$0.37 \pm 0.00$ &$0.14 \pm 0.00$ &$0.23 \pm 0.01$ & $0.16$\\
        & + 90° rot. & $0.23 \pm 0.01$&$0.35 \pm 0.02$&$0.14 \pm 0.01$&$0.25 \pm 0.01$ &$0.13$\\
        & + rand. rot. & $0.21 \pm 0.00$ & $0.37 \pm 0.02$ & $0.16 \pm 0.00$ & $0.26 \pm 0.00$ & $0.06$  \\        
        \midrule        
        \multirow{3}{*}{\tsne} & pixel space & $-0.09$ & $0.07$ & $0.08$ & $-0.05$ &$0.02$ \\        
        & ResNet18 & $-0.11$ & $0.13$ & $0.14$&$0.17$ &$0.04$ \\
        & ResNet152 & $-0.15$ & $0.03$ &$0.14$&$0.19$ &$0.05$\\
    \bottomrule
    \end{tabular}
\end{table}

In this study, we asked (i) how the contrastive visualisation method $t$-SimCNE \citep{boehm2023unsupervised} could be applied to medical image datasets, and (ii) if the set of data augmentations could be enriched compared to what is typically used on natural images.

We considered the Leukemia dataset as our first example (Figure~\ref{fig:leukemia}). Naive application of $t$-SNE to the raw images in pixel space resulted in an embedding with little class separation and low $k$NN accuracy of 67.4\% (Figure~\ref{fig:leukemia}a). Passing all images through an ImageNet-pretrained ResNet and then embedding them with $t$-SNE improved the $k$NN accuracy to 82.2\% but visually the classes were still separated poorly (Figure~\ref{fig:leukemia}b). Training $t$-SimCNE with default data augmentations gave embeddings with 86.7\% $k$NN accuracy (Table~\ref{tab:accuracy}) and much better visual class separation (Figure~\ref{fig:leukemia}c and Table~\ref{tab:silhoutte_score}). This shows that $t$-SimCNE can produce meaningful 2D visualizations of medical image datasets. 

We reasoned that the set of data augmentations could be enriched to include 90° rotations and flips because the semantics of blood microscopy images is rotationally invariant. When training $t$-SimCNE with this set of data augmentations, the $k$NN accuracy increased to 94.0\%. Additionally including all possible rotations by an arbitrary angle as data augmentations yielded the highest $k$NN accuracy (95.2\%) and the highest silhouette score (0.24), indicating that domain-specific augmentations can further improve $t$-SimCNE embeddings.

\begin{figure}[t]
    \centering
    \includegraphics{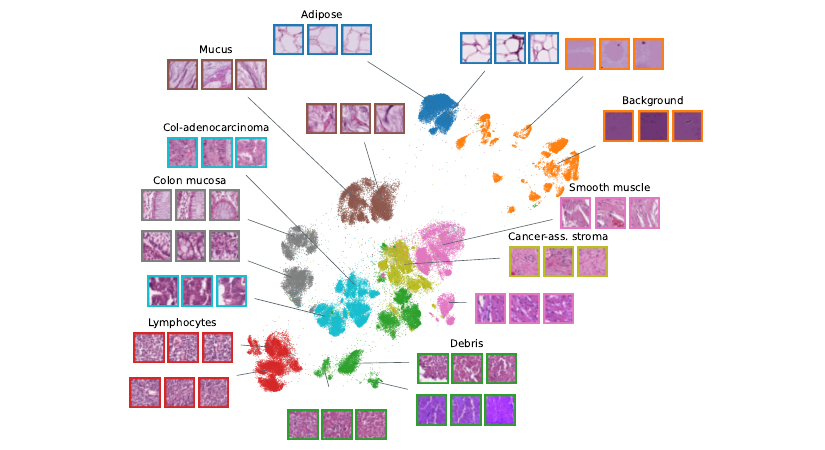}
    \caption{$t$-SimCNE visualisation of the Path{\scshape mnist} dataset. Colours correspond to classes. Images correspond to three random points close to the tip of the annotation line.}
    \label{fig:path_annotation}
\end{figure}

Across the five datasets considered in this study, we saw three different outcomes. On microscropy datasets (Leukemia and Blood{\scshape mnist}), $t$-SimCNE with random rotations performed the best: it had by far the best silhouette score (Table~\ref{tab:silhoutte_score}) and visually the most separated classes (Figure~\ref{fig:f3_toy_annotation}a,b). SimCLR followed by $t$-SNE has also benefited from rotational augmentations. Compared to $t$-SimCNE, it had slightly higher $k$NN accuracies (Figure~\ref{tab:accuracy}), but much lower silhouette scores.

\begin{figure}[t]
    \centering
    \includegraphics{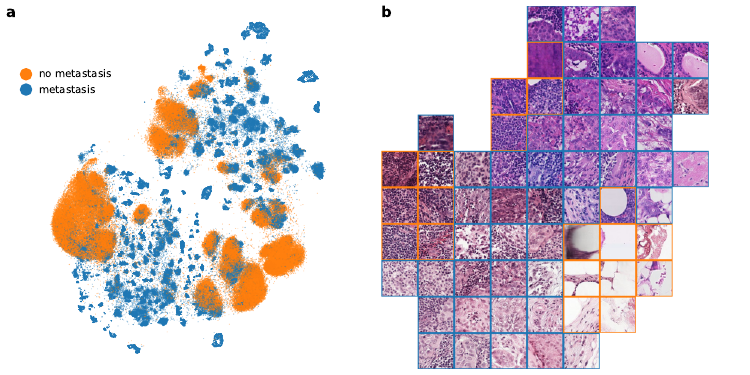}
    \caption{\textbf{(a)} $t$-SimCNE visualisation of the PCam16 dataset. \textbf{(b)} We superimposed a $10\times 10$ grid over the embedding and selected one image in each square. Frame colours show image classes. If a square had fewer than 100 images, no image was shown.}
    \label{fig:camelyon_annotation}
\end{figure}

On pathology datasets (Path{\scshape mnist} and PCam16), $t$-SimCNE with 90° rotations performed the best. On Path{\scshape mnist}, it had the highest silhouette score (Figure~\ref{fig:path_annotation}). On PCam16,  $t$-SimCNE showed clearer structures compared to SimCLR + $t$-SNE, but this difference was not captured by the silhouette scores which on this dataset were all close to zero (Table~\ref{tab:silhoutte_score}). This is because it only had two classes, whereas $t$-SimCNE separated images not only by class but also by tissue types (Figure~\ref{fig:camelyon_annotation}); this led to large within-class distances and hence misleadingly low silhouette scores. 

Finally, on the dermatology dataset (Derma{\scshape mnist}), performance of all methods was similarly poor: SimCLR and $t$-SimCNE resulted in embeddings not very different from $t$-SNE in pixel space (Figure~\ref{fig:f3_toy_annotation}c).

As a control experiment, we applied $t$-SimCNE with  90° rotations and vertical flips to the CIFAR-10 dataset \cite{krizhevsky2009learning}. It decreased the $k$NN accuracy from 89\% to 76\%. This confirms that rotation augmentations are not helpful for natural images because they are not invariant to rotations, unlike microscopy and pathology images.
 
In the pathology datasets, $t$-SimCNE showed meaningful subclass structure. For example, in Path{\scshape mnist}, the \textit{debris} class separated into three clearly distinct subsets (Figures~\ref{fig:path_annotation}), one of which had markedly different staining colour. 
In the PCam16 dataset, the embedding clearly split patches with and without metastasis, based on the density of chromatin and variation in the size of the cells. The difference in visual appearance (different shades of violet) between top-right and bottom-left likely reveals a technical artefact resulting from different staining durations.   

\section{Discussion}

In this paper, we showed that $t$-SimCNE \citep{boehm2023unsupervised} can be successfully applied to medical image datasets, yielding semantically meaningful visualisations, and benefits from rotational data augmentations, leveraging rotational invariance of microscropy images. In agreement with \citet{boehm2023unsupervised}, $t$-SimCNE performed better than SimCLR + $t$-SNE combination. Even though SimCLR tended to have slightly higher $k$NN accuracy, the silhouette score was typically much lower: $t$-SimCNE achieved visually much stronger cluster separation, which is useful for practical visualisations. Furthermore, parametric nature of $t$-SimCNE allows to embed new (out-of-sample) images into an existing embedding.

We found that blood microscopy datasets benefited the most from random rotations, while pathology datasets showed the best results with 90° rotations and flips. We believe it is because in blood microscopy images, the semantically meaningful part is always in the center (Figure~\ref{fig:f3_toy_annotation}a,b) and so the corners of the image may not be important. In contrast, in histopathology images, the edges of the image may contain relevant information, which may get rotated out of the image and replaced by solid-color triangles (Figure~\ref{fig:augm-isolation}c). One of the datasets, Derma{\scshape mnist}, exhibited poor results with all analysis methods. This may be because in this dataset the images are too small to convey biomedically relevant information, or because the sample size was insufficient (Table~\ref{tab:dataset-summary}). 

In conclusion, we argue that $t$-SimCNE is a promising tool for visualisation of medical image datasets. It can be useful for quality control, highlighting artefacts and problems in the data. It can also create a 2D map of cell types, tissue types, or medical conditions, which can be useful not only for clinical purposes but also education and research, potentially combined with an interactive image exploration tool. In the future, it may be interesting to extend $t$-SimCNE to learn representations invariant to technical (e.g. staining) artefacts.

\newpage

\midlacknowledgments{
We thank Christian Schürch for discussion on histopathology data. This work was supported by the German Science Foundation (Excellence Cluster 2064 ``Machine Learning --- New Perspectives for Science'', project number 390727645), the Hertie Foundation, and the Cyber Valley Research Fund (D.30.28739). The authors thank the International Max Planck Research School for Intelligent Systems (IMPRS-IS) for supporting Jan Niklas Böhm. Philipp Berens is a member of the Else Kr\"oner Medical Scientist Kolleg ``ClinbrAIn: Artificial Intelligence for Clinical Brain Research''.
}

\bibliography{references}

\newpage
\appendix
\section*{Appendix} 

\counterwithin{table}{section}
\renewcommand{\thetable}{S\arabic{table}}
\setcounter{table}{0}

\begin{table}[h!]
    \caption{Ablation study, removing individual augmentations from \tsimcne. The full set of augmentations included the default \tsimcne{} augmentations plus arbitrary rotations ($k$NN accuracy is given in percents).}
    \label{tab:ablations}
    \centering\footnotesize
    \begin{tabular}{lrrrrrr}
        \toprule
        Augmentations & \multicolumn{2}{c}{Leukemia} & \multicolumn{2}{c}{Blood{\scshape MNIST}}& \multicolumn{2}{c}{Path{\scshape MNIST}}\\
        \cmidrule(lr){2-3} \cmidrule(lr){4-5} \cmidrule(lr){6-7}
        & $k$NN acc. & Silhouette  & $k$NN acc. & Silhouette & $k$NN acc. & Silhouette  \\
        \midrule
        All & $95.1 \pm 0.2$ & $0.52\pm 0.02$ & $92.9\pm 0.1$ & $0.50 \pm 0.01$ &$97.3 \pm 0.0$&$0.41 \pm 0.03$\\
        No crops & $79.7  \pm 0.6$ & $0.14 \pm 0.00$ & $76.0 \pm 1.1$ & $0.20  \pm 0.01$ & $59.8 \pm 1.1$ &$-0.02 \pm 0.03$\\
        No color jitter & $82.0 \pm 0.2$ & $-0.01 \pm 0.01$ & $90.0 \pm 0.1$ & $0.45 \pm 0.02$ &$94.3 \pm 0.3$&$0.24 \pm 0.02$\\
        No grayscaling & $95.6  \pm 0.4$ & $0.52  \pm 0.02$ & $92.1 	 \pm 0.3$ & $0.44 \pm 0.01$ &$98.5 \pm 0.0$ &$0.39 \pm 0.05$\\
    \bottomrule
    \end{tabular}
\end{table}

\end{document}